\documentclass[final]{cvpr}

\usepackage{times}
\usepackage{epsfig}
\usepackage{graphicx}
\usepackage{amsmath}
\usepackage{amssymb}

\usepackage{subcaption}
\usepackage{flushend}
\usepackage[pagebackref=true,breaklinks=true,colorlinks,bookmarks=false]{hyperref}

\def\cvprPaperID{13} % *** Enter the CVPR Paper ID here
\def\confYear{CVPR 2021}

\begin{document}

%%%%%%%%% TITLE
\title{Color Counting for Fashion, Art, and Design}

\author{Mohammed Al-Rawi\\
ADAPT Centre, Trinity College Dublin\\
Dublin, Ireland\\
{\tt\small alrawim@tcd.ie} }

% \twocolumn[{%
%     % \renewcommand\twocolumn[1][]{#1}%
%     \maketitle
%     \centering
%     \includegraphics[width=.8\textwidth,height=5cm]{example-image} \\
%     \caption{Test caption}
%     \label{teaser}
% }]

\maketitle

%%%%%%%%% ABSTRACT
\begin{abstract}
   Color modelling and extraction is an important topic in fashion, art, and design. Recommender systems, color-based retrieval, decorating, and fashion design can benefit from color extraction tools. Research has shown that modeling color so that it can be automatically analyzed and / or extracted is a difficult task. Unlike machines, color perception, although very subjective, is much simpler for humans. That being said, the first step in color modeling is to estimate the number of colors in the item / object. This is because color models can take advantage of the number of colors as the seed for better modelling, \eg to make color extraction further deterministic. We aim in this work to develop and test models that can count the number of colors of clothing and other items. We propose a novel color counting method based on cumulative color histogram, which stands out among other methods. We compare the method we propose with other methods that utilize exhaustive color search that uses Gaussian Mixture Models (GMMs) and K-Means as bases for scoring the optimal number of colors, in addition to another method that relies on deep learning models. Unfortunately, the GMM, K-Means, and Deep Learning models all fail to accurately capture the number of colors. Our proposed method can provide the color baseline that can be used in AI-based fashion applications, and can also find applications in other areas, for example, interior design. To the best of our knowledge, this work is the first of its kind that addresses the problem of color-counting machine.
   
\end{abstract}

\begin{figure*}
    \begin{center}
    \includegraphics[scale=0.2]{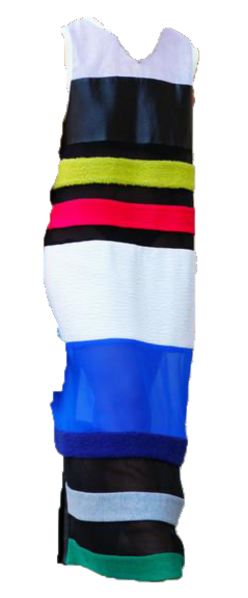} \hspace{1cm}
    \includegraphics[scale=0.3]{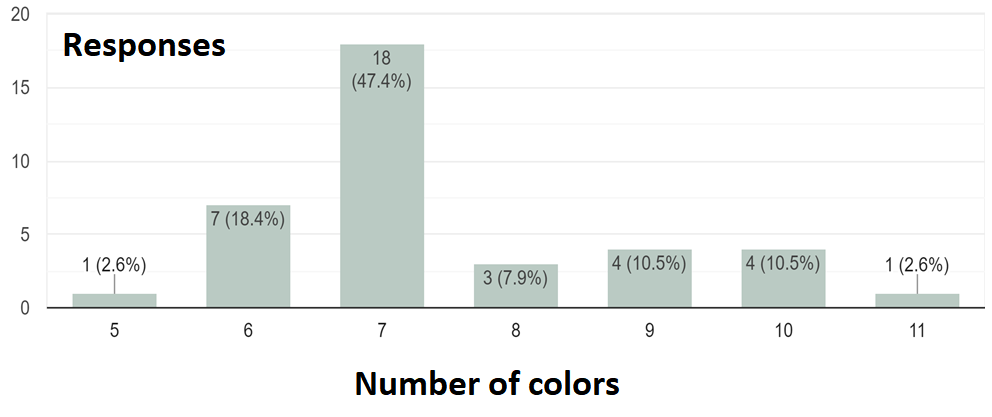}
    \caption{Color count is subjective. The dress has 9 distinct colors, but people have different opinions.}
    \end{center}
    \label{ColorSubjective}
\end{figure*}

%%%%%%%%% BODY TEXT
\section{Introduction}
Automated color extraction is getting more attention in digital art work and design. This includes, but is not limited to, fashion and decoration and vast number of applications, like recommender systems. Digital images are the media that are normally used to mimic real world objects. However, problems of color degradation and the vast number of colors available makes automated color estimation a difficult problem \cite{Hpalette2018,FashCV-srvy20}. The first step to assist models accurately extract colors is knowing the how many colors are there in a scene or object. While this might initially appear to be quite simple, in reality, it is a challenging problem to overcome. In fact, estimating the number of colors in a scene is highly subjective even for humans (even after excluding protans and deutans; \eg people with some form of color blindness).

Color counting is a highly intelligent task as it requires dual cognitive modes; recognizing colors while discarding spatial information, and counting intelligence. While color counting has long been used to teach toddlers, there are still not enough resources nor models to keep machines on par with toddlers color counting and matching skills. Hence, the computer vision research community has devoted their efforts to extracting colors directly from the images, by skipping the color counting step. In this direction, clustering algorithms have widely been used, ~\cite{ACP2005, ACP2007, FCMColor2011, LinCol2013,ZPal2017, Hpalette2018}, such that the number of colors must be known / given a prior. A more recent work proposed a multistage approach to automatically extract the color values; clustering with large number of colors in the first stage followed by merging the clustered colors, based on the hue values, in the second stage ~\cite{Rawi2020}. 

Classification techniques have also been used to categorize colors as in  \cite{Manfredi2013ACS, Yu28Color, CVC2018}. Classification techniques have the disadvantage that they are limited to a predefined number of tagged colors, and worse, the problem of memorizing the colors and shapes they see at training (\ie the generalization dilemma). Moreover, treating color modeling as a classification will lead to failure of keeping pace with the millions of colors used in images  \cite{Deane75}.

This work addresses the "color counting" problem and thus differs from previous methods that jump straight into color extraction. By knowing the exact number of colors, one can easily extract colors from images, as this will make color extraction deterministic. 
%---------------------------------------------------------------------

\section{Methods}
\section{Color Distribution}

Color images usually suffer from severe distortions distortions. Color printing quality, color interlacing, photographic geometry, amount of light, image compression, and even imaging devices affect how colors appear in images and the amount of noise they may contain. Al-Rawi and Joeran showed in \cite{Rawi2020} that the colors of a multichnnel (i.e., RGB) image can be modeled as a Gaussian Mixture Model (GMM) prior distribution, which is given by:

\begin{equation}
p(\mathbf{x})=\sum _{i=1}^{K}\phi _{i}\mathcal{N}(\mathbf{x}|\text{\boldmath$\mu$}_{i}, \mathbf{\Sigma_i}),
\end{equation}
where
\begin{multline*}
      \mathcal{N}(\mathbf{x}|\text{\boldmath$\mu$}_{i}, \mathbf{\Sigma_i}) = \\
      \frac{1}{\sqrt{(2\pi)^K|\mathbf{\Sigma_i}|}} \exp{\Big(-(\mathbf{x}-\text{\boldmath$\mu$}_{i})^\text{T}\mathbf{\Sigma_i}^{-1}(\mathbf{x}-\text{\boldmath$\mu$}_{i}) \Big)},
  \label{GMM}
 \end{multline*}
 
\noindent and $K$ denotes the number of colors, and the $i_\text{th}$ vector component is characterized by normal distributions with weight ${\displaystyle \phi _{i}}$, mean 
${\boldsymbol {\mu _{i}}}$ and covariance matrix ${\boldsymbol {\Sigma _{i}}}$. The color distribution becomes extremely complicated when the clothing item has more than one color and because of the possibility of frequency deviation of colors during the imaging process. In this case, the GMM prior distribution will be given by:
\begin{equation}
p(\mathbf{x})=\sum _{i=1}^{K'}\phi _{i}\mathcal{N}(\mathbf{x}|\text{\boldmath$\mu$}_{i}, \mathbf{\Sigma_i})),
\label{GMM2}
\end{equation}
where $K'$ denotes the total number of colors, or model components, such that $K'=K+K_f$, and $K_f$ denotes the number of new (but fake) colors generated during image acquisition. To simplify notations and without loosing generality, we deffer in what comes below from using the $K'$ notation and use $K$ instead. The GMM is also disrupted due to the use of 8 bits for each pixel in the image, as it leads to truncating the pixel values close to the lower-bound and / or upper-bound, \textit{i.e.} values close to 0 or 255 for an 8-bit per pixel image. Therefore, there will always be some form of incorrectness in color distributions in real-world images containing values close to these extreme values; the so we call truncated tail effect. 

\subsection{Color counting via GMMs}
As aforementioned, GMMs denote the mathematical, and natural, distribution of colors in images. This suggests that one can model colors as a GMM to estimate the correct number of colors. The hypothesis is that the optimal number of colors, $K$, yields to the optimal 1) score value computed as the per-sample average log-likelihood of the given color values; 2) Akaike’s information criterion (AIC); 3) the Bayesian information criterion (BIC), and 4) the Jensen-Shannon distance (JS-distance) between two GMMs randomly sampled from the given data. Testing these hypotheses requires running exhaustive search processes up to a pre-defined maximum number of colors in the image; \ie by varying $K$ from 1 to $K_{max}$. The optimal number of components will correspond to some extreme values of AIC or BIC \cite{GMM2014b}, or JS distance. This is widely known in the multivariate literature as estimating the number of components \cite{GMM2014}. However, while the literature shows some success in using these scores to estimate the number of components, these studies mostly rely on simulated data \cite{GMM2014b}; opposed to real data that has some form of distortion and non-trivial noise, as in the color distribution case. Although the GMMs computational burden is very high, obtaining a correct color count via GMMs would be highly beneficial. We shall refer to this method as CC-GMM.

\newcommand{\sz}{0.35}
\begin{figure*}[!htp]
    
    \centering
     \subcaptionbox{multi bar, \#colors=16(24)}{\includegraphics[scale=\sz]{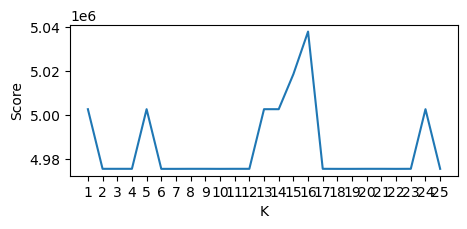}}
     \subcaptionbox{8 bars, \#colors=4(8)}{\includegraphics[scale=\sz]{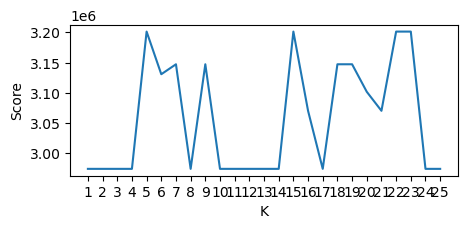}}
     \subcaptionbox{chair, \#colors=9(4)}{\includegraphics[scale=\sz]{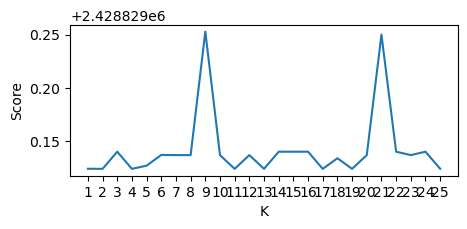}}
     \subcaptionbox{waves, \#colors=15(24)}{\includegraphics[scale=\sz]{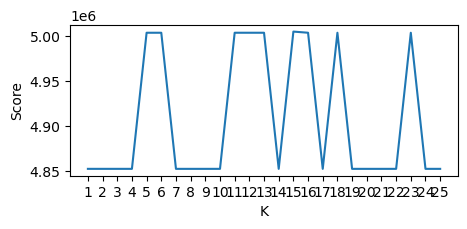}}
     \vspace{.5cm}

     \subcaptionbox{jacket, \#colors=5(1)}{\includegraphics[scale=\sz]{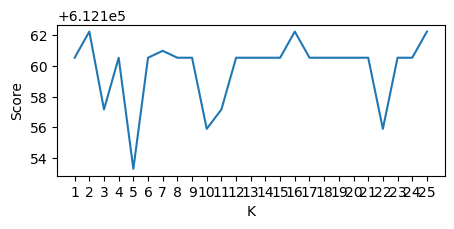}}
     \subcaptionbox{dress, \#colors=9(9)}{\includegraphics[scale=\sz]{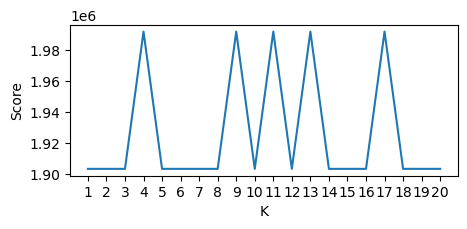}}
     \subcaptionbox{white-shirt, \#colors=19(1)}{\includegraphics[scale=\sz]{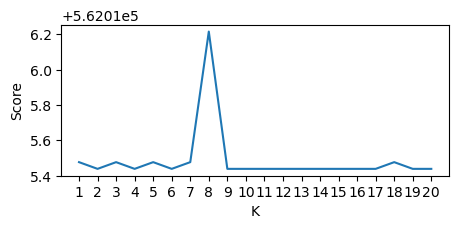}}
     \subcaptionbox{sweatshirt, \#colors=18(4)}{\includegraphics[scale=\sz]{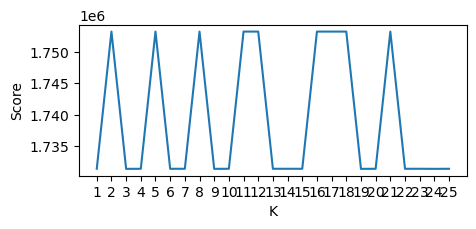}}
     
    % I don't have enough space, so I am removing this figure 
    %  \vspace{.5cm}

    %  \subcaptionbox{pants, \#colors=9(1)}{\includegraphics[scale=\sz]{AicFigures/pants_gmm_aic_Kfound9_21s.png}}

    \caption{Color counting using Gaussian Mixture Model, CC-GMM. Score shown on the y-axis denotes the AIC (Akaike’s information criterion). We estimated the GMM score over K={1, 2,..., 25}. The sub-caption  \#colors=$x$($T$) denotes the number of colors, where $x$ is estimated and ($T$) is the ground-truth. As GMMs are stochastic, running the experiments again will lead to different results each time.}
    
    \label{CC-GMM}
\end{figure*}

\newcommand{\hsp}{\hspace*{.2cm}}
\newcommand{\hspk}{\hspace*{.4cm}}
\newcommand{\szf}{0.2}

\begin{figure*}[ht]
    
    \centering
     \subcaptionbox{multi bar}{\includegraphics[scale=\szf]{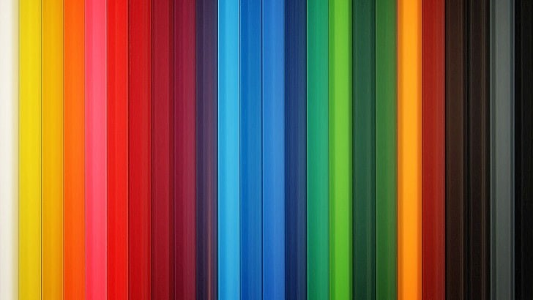}}
     \subcaptionbox{\#colors=19(24)}{\includegraphics[scale=\szf]{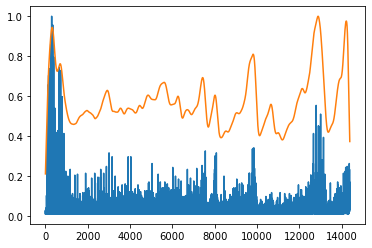}} \hsp
     \subcaptionbox{8 bars}{\includegraphics[scale=0.12]{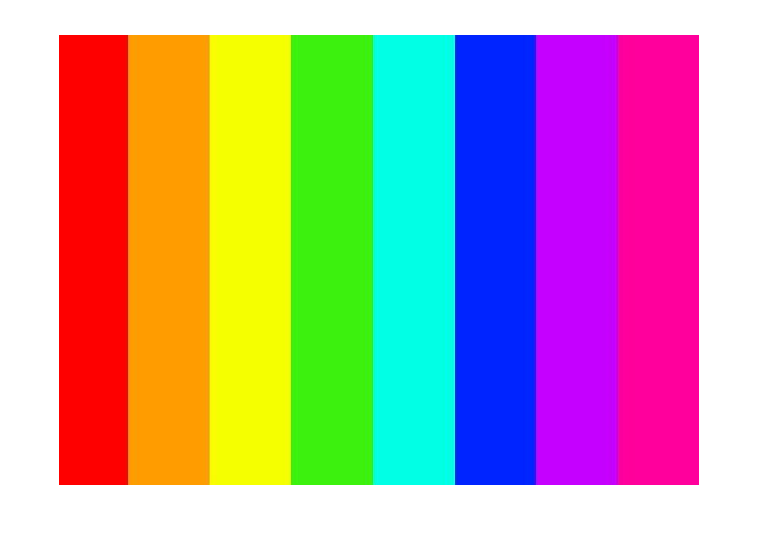}}
     \subcaptionbox{\#colors=6(8)}{\includegraphics[scale=\szf]{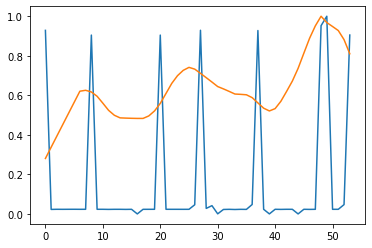}} \hsp
     \subcaptionbox{chair}{\includegraphics[scale=0.05]{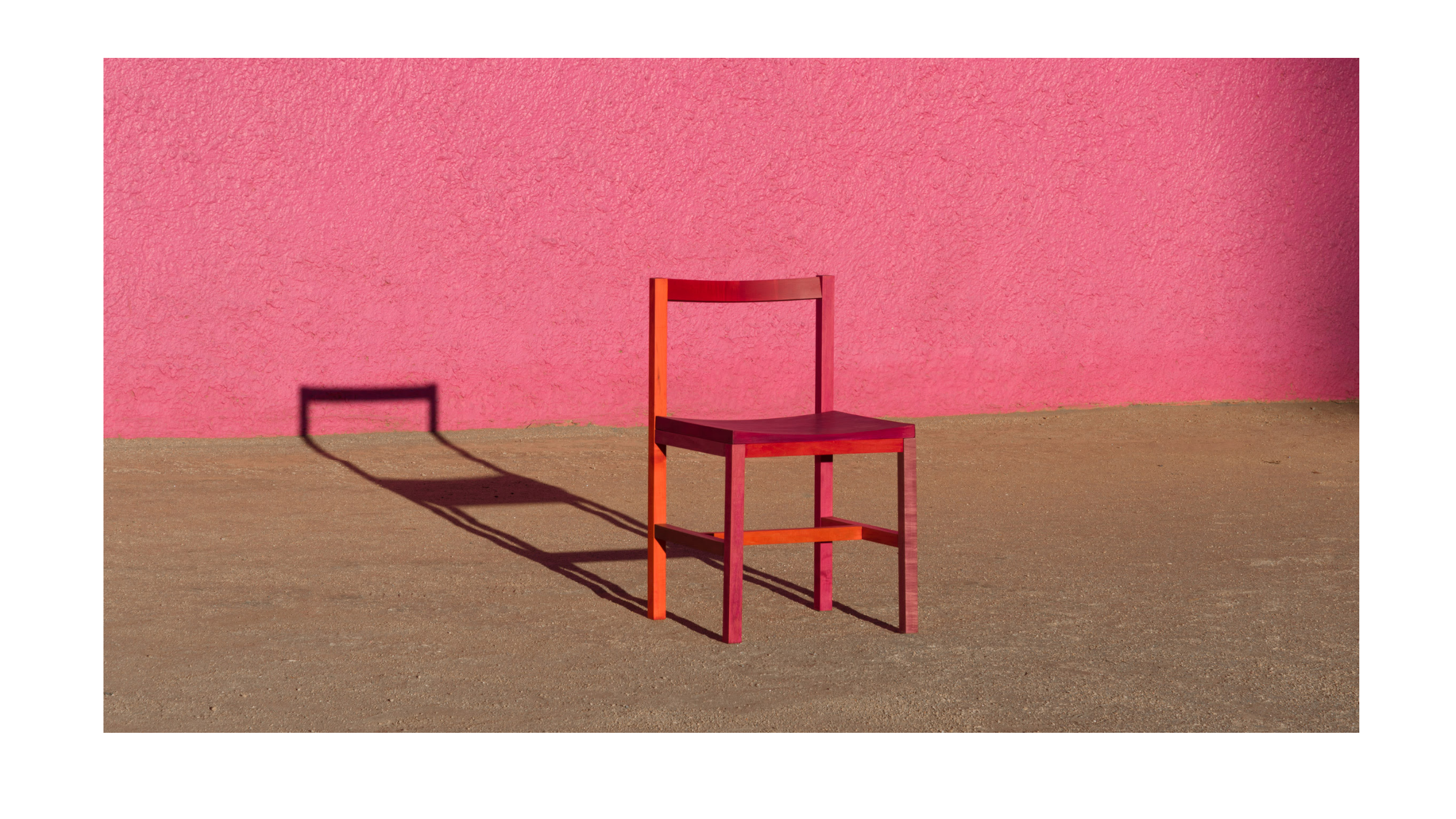}} 
     \subcaptionbox{\#colors=4(4)}{\includegraphics[scale=\szf]{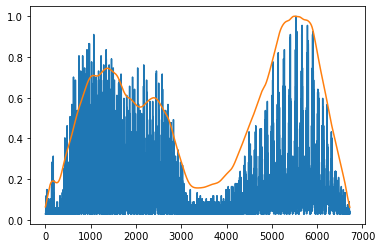}}  \\ \vspace{.5cm}
     
     \subcaptionbox{waves}{\includegraphics[scale=0.2]{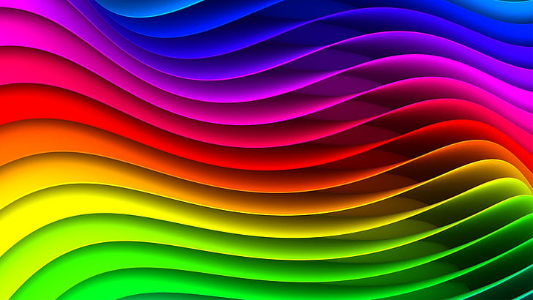}}
     \subcaptionbox{\#colors=19(24)}{\includegraphics[scale=\szf]{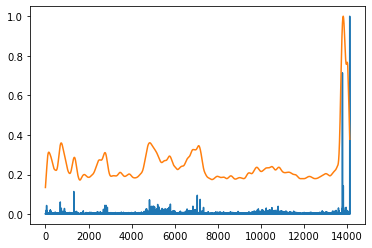}} \hspk
     \subcaptionbox{jacket}{\includegraphics[scale=\szf]{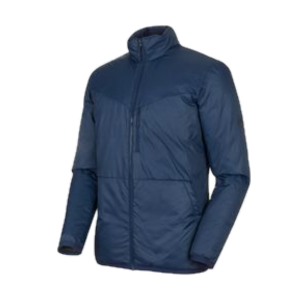}}
     \subcaptionbox{\#colors=1(1)}{\includegraphics[scale=\szf]{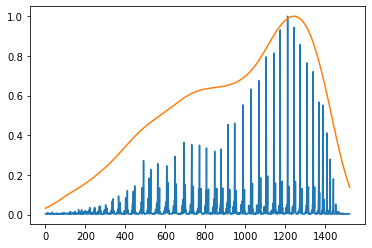}} \hspk
     \subcaptionbox{shirt}{\includegraphics[scale=\szf]{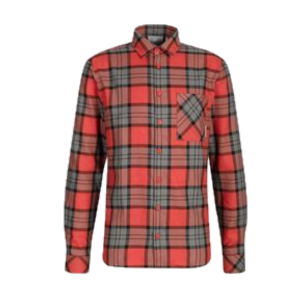}}
     \subcaptionbox{\#colors=4(4)}{\includegraphics[scale=\szf]{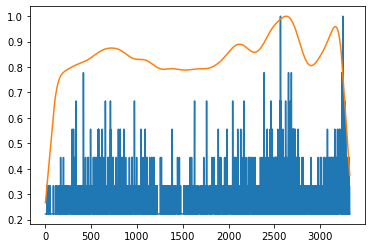}} \\ \vspace{.5cm}
     
     \subcaptionbox{pants}{\includegraphics[scale=\szf]{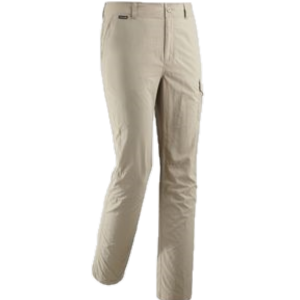}}
     \subcaptionbox{\#colors=1(1)}{\includegraphics[scale=\szf]{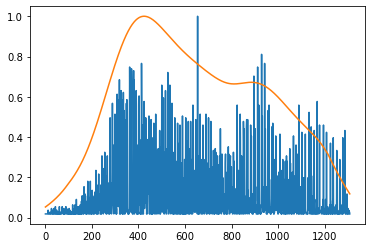}} \hspk     
     \subcaptionbox{dress}{\includegraphics[scale=0.13]{figures/dress.png}}
     \subcaptionbox{\#colors=8(9)}{\includegraphics[scale=\szf]{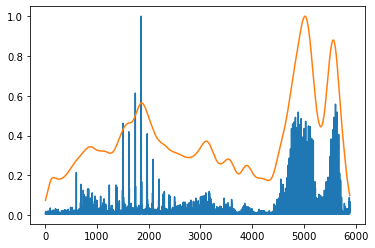}} \hspk     \subcaptionbox{sweatshirt}{\includegraphics[scale=0.16]{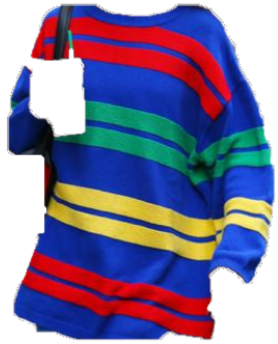}}
     \subcaptionbox{\#colors=5(4)}{\includegraphics[scale=\szf]{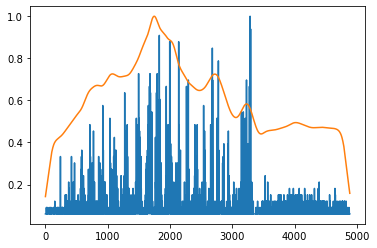}} \hspk 
     \subcaptionbox{white-shirt}{\includegraphics[scale=0.23]{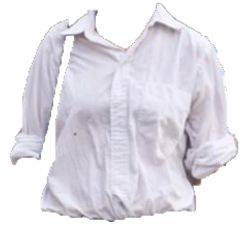}}
     \subcaptionbox{\#colors=2(1)}{\includegraphics[scale=\szf]{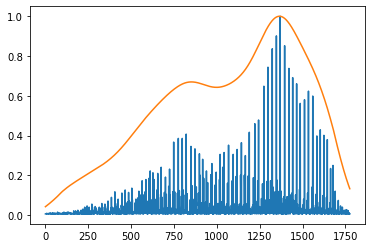}}
     
    \caption{Color counting using our proposed method, CC-CH. For each consecutive pair (\eg, a and b), we have the image in (a) and its distribution in (b). The sub-caption  \#colors=$x$($T$) denotes the number of colors, where $x$ is estimated and ($T$) is the ground-truth. We scaled the images to fit in the figure; that is, they are of different resolutions. Best viewed in color.}
    
    \label{CC-CH}
\end{figure*}

\subsection{Color counting via K-Menas}
It has been shown in \cite{kmGauss} that K-Means clustering can be used to approximate a GMM. Although K-Means clustering is computationally problematic (it is in fact NP-hard), efficient heuristic algorithms quickly converge to a local minima. We will use K-Means clustering as another method to estimate the number of colors in images. We use the opposite of the value of the data (image) on the K-means objective function. We shall refer to this method as CC-KM.

\subsection{Color counting via deep convolutional neural networks}
We use self-supervised learning to train a deep convolutional neural network (DCNN) for it to work as a color counting model. The DCNN is a classifier, and the self learning is done by randomly generating shapes of clothing and other items that have different colors and shapes. The generator passes the image and the number of colors to the DCNN. We use EfficientNet in PyTorch to model the DCNN. We shall refer to this method as CC-DCNN.

\subsection{Color counting via cumulative histogram}
In this method, we calculate the histogram of each band of the RGB image. Then, we arrange the histogram as a triplet vector that depends on the extreme values in each channel. We further compress the histogram by removing the outliers, which we define by the use of Principle Component Analysis (PCA). Our hypothesis in this method is that colored pixels in the image will produce peaks in the cumulative triplet vector which we compute as mentioned above. Removal of outliers is an important step that controls the color count, aimed at removing a few peaks resulting from noise and distortion. The number of peaks will therefore give an estimate of the number of colors in the image. We shall refer to this method as CC-CH

\begin{figure}[ht]
    
    \centering
     %\subcaptionbox{multi bar}
     {\includegraphics[scale=0.5]{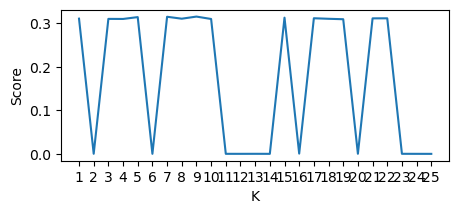}}
     
    \caption{Score denotes the JS-distance between two GMMs calculated for the sweatshirt image.}
    
    \label{JS-dist}
\end{figure}

\section{Results}
We collected a benchmark of 100 different color images of fashion and art articles that have a distinctive number of colors. We use sklearn to model GMMs, and PyTorch for DCNNs. Due to lack of space, we visually demonstrate experiments selected from the results that conform with whole dataset. We firs use the K-Means algorithm from the sklearn package. The values of K ranges from 1 to 25. The K-Means algorithm, although faster than the GMM, is unsuccessful in estimating the number of colors. This is because the score curve lacks distinctive extreme values that identify the number of colors. In addition, the DCNN has also been unsuccessful in counting the colors. While the DCNN converged well on the training set, it failed on the testing set; which means it has serious overfitting issues due to memorizing color features and focusing on the spatial information (shape) and not the targeted objective, which is color count. The GMM approach results in slightly better performance than the two aforementioned methods, yet, it is still far from being accurate as shown in Fig. \ref{CC-GMM}. Moreover, GMMs are also stochastic by nature; for example, over 10 runs on the multi-bar image shown in Fig. \ref{CC-CH}, estimated color counts are \{15, 11, 6, 16, 24, 4, 7, 10, 14, 5\}. We present in Fig. \ref{JS-dist} the JS-distance between two GMMs for the sweatshirt image, which has four colors. We randomly split the image data into two halves, each used to estimate one GMM, then we sample 100000 samples from each GMM to calculate the distance. The curve has no distinctive behaviour, and the minimum distance is found at K=24. Further experiments with other images are consistent with this conclusion. This indicates that the JS-distance is not effective to estimate the number of colors. There are some other methods that we have tested that also were unsuccessful, and the discussion goes beyond the limit of this work; such as Bayesian Gaussian Mixture Models, Siamese DCNNs, mean of GMMs / K-Menas distributions over several iterations, and probing changes in clusters' sizes. Our proposed method, CC-CH, emerged after exploring all these aforementioned approaches. Results depicted in Fig. \ref{CC-CH} show CC-CH is very promising . We can see that it can result in a correct number of colors, or a number close to the correct number of colors in the image.

\section{Conclusion}
We proposed a novel method to count the number of colors in RGB images. Our cumulative histogram method is not only more accurate than the GMM, but also 100 folds faster. Over the benchmark we used, the average execution time of the GMM is ~100 seconds, while the method we propose needed less than ~1 second. We implemented the experiments on the CPU and we can still have more reduction in execution time if we are to use the GPU. Furthermore, our proposed method is deterministic and does not depend on some random initialization parameters. That being said, one gets exactly the same answer at every run. The GMMs, on the other hand, need to randomly initialize the parameters of their components. Therefore, GMMs are highly stochastic and one might not be able to replicate the results or even ensure the correct ones. Moreover, our experiences show that color counting via GMMs, although slightly better than K-Means and DCNNs, is incorrect and unreliable.

\section*{Acknowledgement}
\noindent\small{This research was conducted with the financial support of European Union’s Horizon 2020 programme under the  Marie Skłodowska-Curie Gran Grant Agreement \# 801522 at the ADAPT SFI Research Centre at Trinity College Dublin. The ADAPT SFI Centre for Digital Content Technology is funded by Science Foundation Ireland through the SFI Research Centres Programme and is co-funded under the European Regional Development Fund (ERDF) through Grant \# 13/RC/2106\_P2. This work was also supported by TCHPC (Research IT, Trinity College Dublin).}

{\small
\bibliographystyle{ieee_fullname}
\bibliography{egbib.bib}
}

\end{document}